\documentclass[10pt,twocolumn,letterpaper]{article}

\usepackage{cvpr}
\usepackage{epsfig}
\usepackage{graphicx}
\usepackage{amssymb}
\usepackage{arydshln}
\usepackage{capt-of}
\usepackage{xcolor}
\usepackage{booktabs}
\usepackage{varwidth}

\usepackage[pagebackref=true,breaklinks=true,letterpaper=true,colorlinks,bookmarks=false]{hyperref}

 \cvprfinalcopy 

\title{BusyHands: A Hand-Tool Interaction Database for \\  Assembly Tasks Semantic Segmentation} 
	
\author{Roy Shilkrot, Zhi Chai and Minh Hoai\\
	Stony Brook University\\
	100 Nicolls rd., Stony Brook, NY 11794\\
	{\tt\small \{roys,zhchai,minhhoai\}@cs.stonybrook.edu}
}

\begin{document}
	
\twocolumn[{%
	\renewcommand\twocolumn[1][]{#1}%
	\maketitle
	\begin{center}
		\centering
		\includegraphics[width=\linewidth]{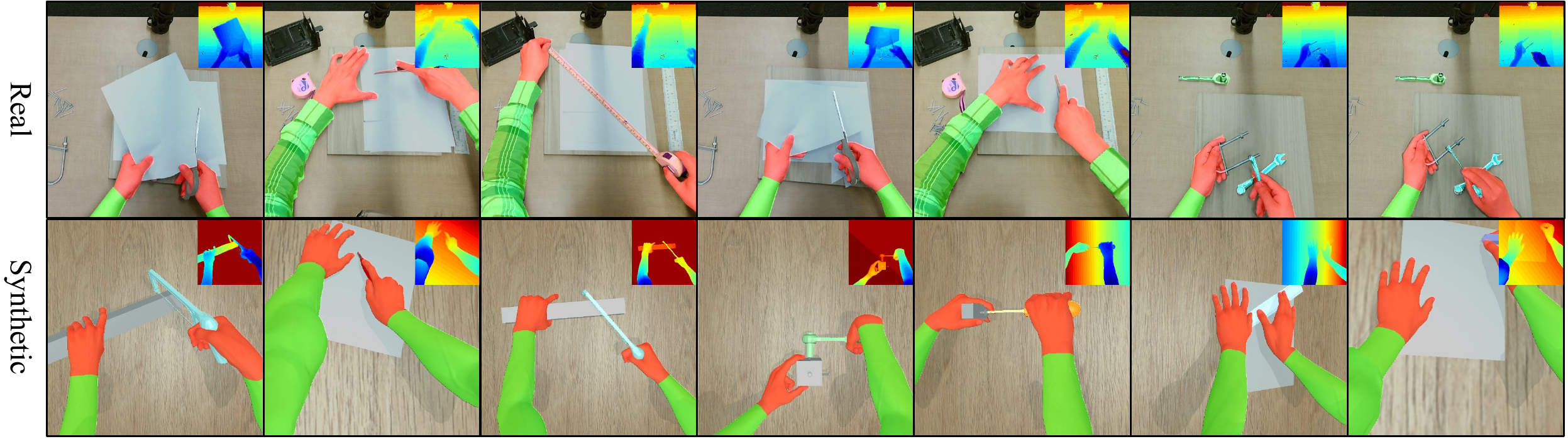}
		\captionof{figure}{A visualization of some annotated samples from our dataset, classification overlaid. We contribute an RGB+D dataset with 16 classes, featuring synthetic and real-world captured data, manually as well as automatically annotated.\label{fig:samples}}
	\end{center}%
}
]

\begin{abstract}
Visual segmentation has seen tremendous advancement recently with ready solutions for a wide variety of scene types, including human hands and other body parts. 
However, focus on segmentation of human hands while performing complex tasks, such as manual assembly, is still severely lacking. 
Segmenting hands from tools, work pieces, background and other body parts is extremely difficult because of self-occlusions and intricate hand grips and poses. 
In this paper we introduce BusyHands, a large open dataset of pixel-level annotated images of hands performing 13 different tool-based assembly tasks, from both real-world captures and virtual-world renderings. 
A total of 7906 samples are included in our first-in-kind dataset, with both RGB and depth images as obtained from a Kinect V2 camera and Blender. 
We evaluate several state-of-the-art semantic segmentation methods on our dataset as a proposed performance benchmark.
\end{abstract}

{\let\thefootnote\relax\footnote{{Dataset link: \href{http://hi.cs.stonybrook.edu/busyhands}{http://hi.cs.stonybrook.edu/busyhands}}}}

\vfill

\section{Introduction}
\begin{quotation}
``\textit{Idle hands} are the devil's playthings''
  --- Benjamin Franklin
 \end{quotation}
Computer vision is now used in many of the manufacturing and fabrication fields. 
Manufacturers are using high-end machine vision for part inspection and verification, as well as means to track the workers and the work pieces to gain crucial insight into the efficiency of their assembly lines.
Small-scale fabrication, on the other hand, happens virtually anywhere, even at home, at school, or in personal fabrication shops.
Still all kinds of fabrication, mass- or small-scale, share a commonality - \textit{manual assembly tasks performed by humans}.
This comes as a stark contrast to the minor offering of computer vision methods to understand manual assembly scenes.
To this end we offer a first-of-its-kind dataset of fully annotated images of assembly tasks with manual tools - named \textit{BusyHands}.
The first offering, described in this paper, includes both real-world and virtual-world samples for semantic segmentation tasks. 
Later iterations of BusyHands will include arm and hand articulated poses (skeleton) as well as multi-part tool 6DOF pose.
We believe an open dataset, such as our BusyHands, can drive research into deeper understanding of manual assembly task imaging, which will in turn help increase efficiency and error-tolerance in industrial pipelines or at home.

Semantic segmentation -- finding contiguous areas in the image with a similar semantic context -- is one of the most fundamental tasks in scene understanding. 
Using a segmentation over the image, further break-down of the parts to smaller parts or interaction between parts can proceed. 
There are numerous popular large-scale standard datasets to assist in segmentation algorithm development, e.g. ImageNet~\cite{JWRL09}, COCO~\cite{LMBB14}, SUN~\cite{XHEO10}, PASCAL~\cite{EVWW10}, and ADE20K~\cite{ZZPF17}.
Further, hand image analysis  datasets~\cite{MZT11,SFMA17,ZB17,M17,AA18} were proposed for segmentation, with a focus on hands, but not hand interactions. Bambach et al.~\cite{BSLS15} create a dataset for complex interactions, but doesn't involve handheld tools. 
Therefore, we find most existing open collections unsuitable for interactions between hands and handheld tools, which is essential for understanding assembly. 

\setlength{\tabcolsep}{3pt}
\begin{table}[t]
	\begin{tabular}{lccc}
		\toprule
		\textbf{Name} & \textbf{\# frames} & \textbf{Depth} & \textbf{Method} \\
		\hline
		EgoHands~\cite{BSLS15} & 4,800 & No & Manual \\
		Handseg~\cite{SFMA17} & 210,000 & Yes & Automatic \\
		NYUHands~\cite{TSLP14} &  6,736 & Yes & Automatic \\
		\cite{ZB17} & 43,986 & Yes & Synthetic \\
		HandNet~\cite{wetzler2015rule} & 212,928 & Yes & Automatic \\
		GTEA~\cite{LYR15} & 663 & No & Manual \\
		\cite{AA18} & 1,590 & No & Manual \\
		\hdashline
		\textbf{Ours} & \textbf{7,905} & \textbf{Yes} & \textbf{Man. \& Syn.}
		\\
		 \bottomrule
	\end{tabular}
	\vspace{0pt}
	\caption{Comparison of hand segmentation datasets. `Depth' indicates the offering of an aligned depth image per RGB image.}
	\label{tab:hand-seg-ds}
\end{table}

\newcommand{\mkhd}[2]{\begin{varwidth}{0.2\linewidth}\centering #1 \\ #2\end{varwidth}}
\begin{table}
		\small
	\centering
	\begin{tabular}{lcccc}
		\toprule
	\textbf{Tool} & \mkhd{\textbf{COCO}}{\cite{LMBB14}} & \mkhd{\textbf{SUN}}{\cite{XHEO10}} & \mkhd{\textbf{ADE20K}}{\cite{ZZPF17}} & \mkhd{\textbf{BusyHands}}{(Ours)} \\

	\hline
Screwdriver	& 0	& 1	& 2	& \textbf{1616}
\\
Wrench	& 0	& 64	& 2	& \textbf{2051}
\\
Pliers	& 0	& 1	& 1	& \textbf{1586}
\\
Pencil	& 0	& 7	& 8	& \textbf{2320}
\\
Scissors & \textbf{975}	& 3	& 16	& 864 \\
Cutter$^{*}$	& \textbf{4507}	& 63	& 161	& 2021
\\
Hammer	& 0	& 4	& 5	& \textbf{1066}
\\
Ratchet	& 0	& 0	& 0	& \textbf{967}
\\
Tape	& 0	& 0	& 1	& \textbf{796}
\\
Saw	& 0	& 1	& 1	& \textbf{1183}
\\
Eraser	& 0	& 0	& 0	& \textbf{846}
\\
Glue	& 0	& 0	& 0	& \textbf{650}
\\
Ruler	& 0	& 2	& 4	& \textbf{2428} \\
\bottomrule
	\end{tabular}
	\vspace{3pt}
	\caption{Comparison of number of pixel-level annotated object instances among prominent segmentation datasets and our own. $^*$~``Knife'' also considered as ``Cutter'' in other datasets.}
	\label{tab:instances-compare}
\end{table}

\begin{figure}[t]
	\includegraphics[width=\linewidth]{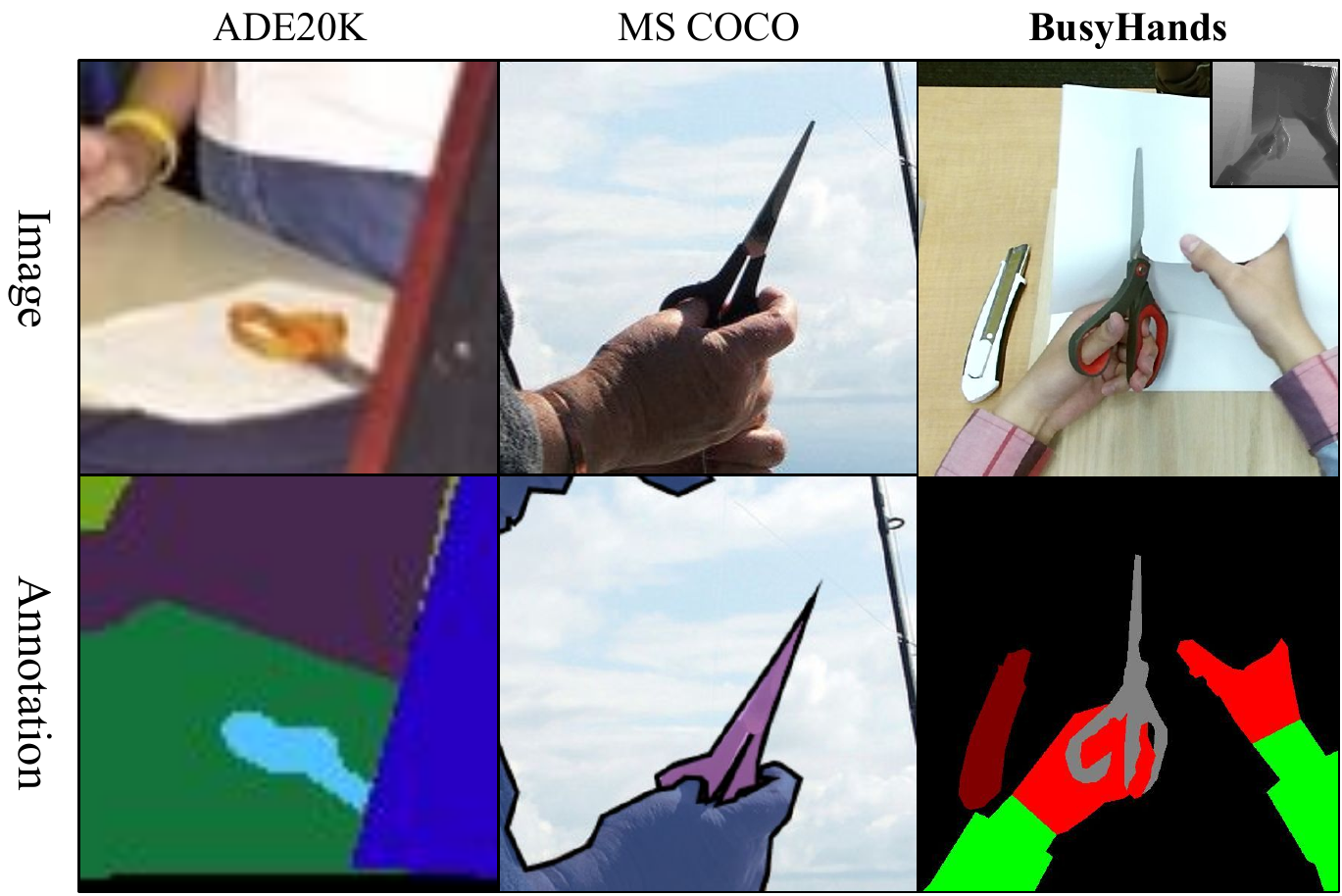}
	\caption{Qualitative comparison of annotation quality in our dataset vs. ADE20K~\cite{ZZPF17} and MS COCO~\cite{LMBB14}. Our annotation is more precise in terms of polygon quality, and the dataset also contains depth information. Additionally, other datasets have a far smaller amount of instances in most object categories (see Table~\ref{tab:instances-compare}).}
	\label{fig:compare-anno}
\end{figure}

Naive methods for human hand segmentation from backgrounds, such as recognizing skin-colored pixels in RGB, are being replaced with supervised machine learning algorithms with far higher perception capabilities, such as deep convolutional networks or deep randomized decision forests.
The advent of new cheap imaging technology, such as the Kinect~\cite{Kinect} depth camera, allowed enriching the fundamental features used in perception tasks to reach (and even surpass) human-level cognitive capabilities.
However, adding more feature dimensions to these highly parametric models requires orders of magnitude more training data to achieve generalizable results. 
Consequently this lead to the construction of the aforementioned large annotated datasets and others, which are now in hard demand.

\begin{figure*}[t]
	\centering
	\includegraphics[width=\linewidth]{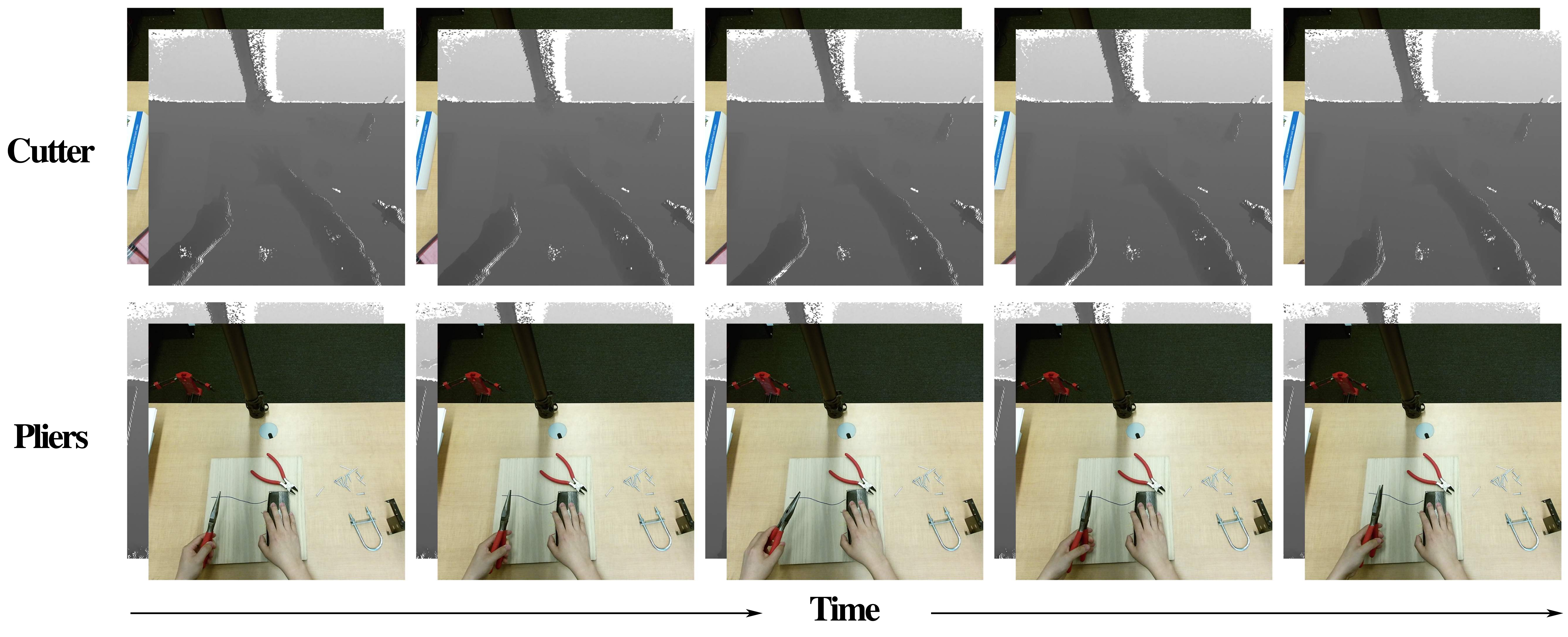}
	\caption{Frame by frame outputs captured from Kinect V2, both color and depth frames are depicted.}
	\label{fig:real_data}
\end{figure*}

Manually annotating distinct semantic parts in images is tedious and error-prone, and therefore it may be prohibitively expensive. 
To cope with this problem, ~\cite{RFRB15,SJMS17} adopted synthetic data which can be generated through professional 3D modeling software. 
Ground truth annotation for semantic segmentation can be achieved easily in 3D software, since the objects are precisely defined (by a triangulated mesh) and photorealistic rendering is ready at hand. 
A 3D model can also be parameterized to augment the data with a multitude of novel situations and camera angles.
Conversely, synthetic scenes also need careful human staging to achieve realism that can generalize to successful real-world data analysis.
All tolled, synthetic datasets are now an advancing reality for many vision tasks, especially in the autonomous driving domain~\cite{RFRB15,SJMS17}.
Therefore, we created BusyHands to have both real-world captures as well as synthetic renderings using Blender. 
We provide a comparative evaluation between real-world and synthetic parts in this paper. 

To the best of our knowledge, ours is the first real- or virtual-world segmentation dataset that focuses on small-scale assembly works. 
A small sample of our annotated dataset is presented in Fig.\ref{fig:samples}. 
We will release for open download all parts of our dataset, as well as all pre-trained segmentation models (see \S\ref{sec:methods}).
A small excerpt from the dataset exists in the supplementary material.

\begin{table*}[t]
	\centering
		\begin{tabular}{ccc}
			\toprule
			& \textbf{Advantage} & \textbf{Disadvantage} \\ \hline
			\parbox{27pt}{\textbf{\\ Real\\ Data}} & 
			\begin{minipage} [t] {0.45\textwidth} 
				\renewcommand{\labelitemii}{$\checkmark$}
				\begin{itemize}\setlength\itemsep{-2pt}
					\item[\checkmark] \textit{Simple to collect data} with commodity cameras.
					\item[\checkmark] Data is \textit{as close as possible to the target} input, thus more attractive to external practitioners.
					\item[\checkmark] Image \textit{capture is immediate}.
					\item[\checkmark] High \textit{data randomness}, assists in generalization.
				\end{itemize} 
			\end{minipage}
			&
			\begin{minipage} [t] {0.45\textwidth} 
				\begin{itemize}\setlength\itemsep{-2pt}
					\item[$\times$] \textit{Annotating is expensive} in terms of time and resources.
					\item[$\times$] Objects might not be labeled correctly due to \textit{occlusion or ambiguity}.
					\item[$\times$] Segmentation may be \textit{subjective}, because of a single annotator or disagreement.
					\item[$\times$] RGB-Depth \textit{registration has artifacts}.\vspace{3pt}
				\end{itemize}
			\end{minipage}
			\\ \hline
			\parbox{20pt}{\textbf{\\ Synth. \\ Data}} &
			\begin{minipage} [t] {0.45\textwidth} 
				\vspace{-8pt}
				\begin{itemize}\setlength\itemsep{-2pt}
					\item[\checkmark] All the images are \textit{annotated accurately and instantly} in an automatic manner. 
					\item[\checkmark] The dataset can be \textit{easily grown} by adding more texture, pose or camera variables.
					\item[\checkmark] RGB and Depth streams are \textit{perfectly aligned}, from the virtual camera's z-buffer.
				\end{itemize} 
			\end{minipage}
			&
			\begin{minipage} [t] {0.45\textwidth} 
				\vspace{-8pt}
				\begin{itemize}\setlength\itemsep{-2pt}
					\item[$\times$] The creation of \textit{3D models and scene} staging is difficult in the earlier stage. 
					\item[$\times$] Realistic \textit{animatronics} is hard to achieve without expertise and resources.
					\item[$\times$] The synthetic images are \textit{not as realistic} as real images, lack noise.
					\item[$\times$] \textit{Image rendering} at high resolution and multiple passes (RGB, Depth map) is time consuming.
				\end{itemize}
			\end{minipage} \\
			\bottomrule
		\end{tabular}
	\vspace{0pt}
	\caption{Advantage analysis of real vs. synthetic data.}
	\label{tab:pros-and-cons}
\end{table*}

The rest of the paper is organized as follows. In Section \ref{sec:related}, we discuss semantic segmentation and existing datasets in the literature. Section \ref{sec:dataset} provides details on how we cerated the BusyHands dataset. Section \ref{sec:evaluation}, covers existing semantic segmentation methods which we used for evaluation on our dataset. Section \ref{sec:conclusion} offers conclusions about this work and future directions.

\begin{figure*}[t]
	\centering
	\includegraphics[width=0.47\linewidth]{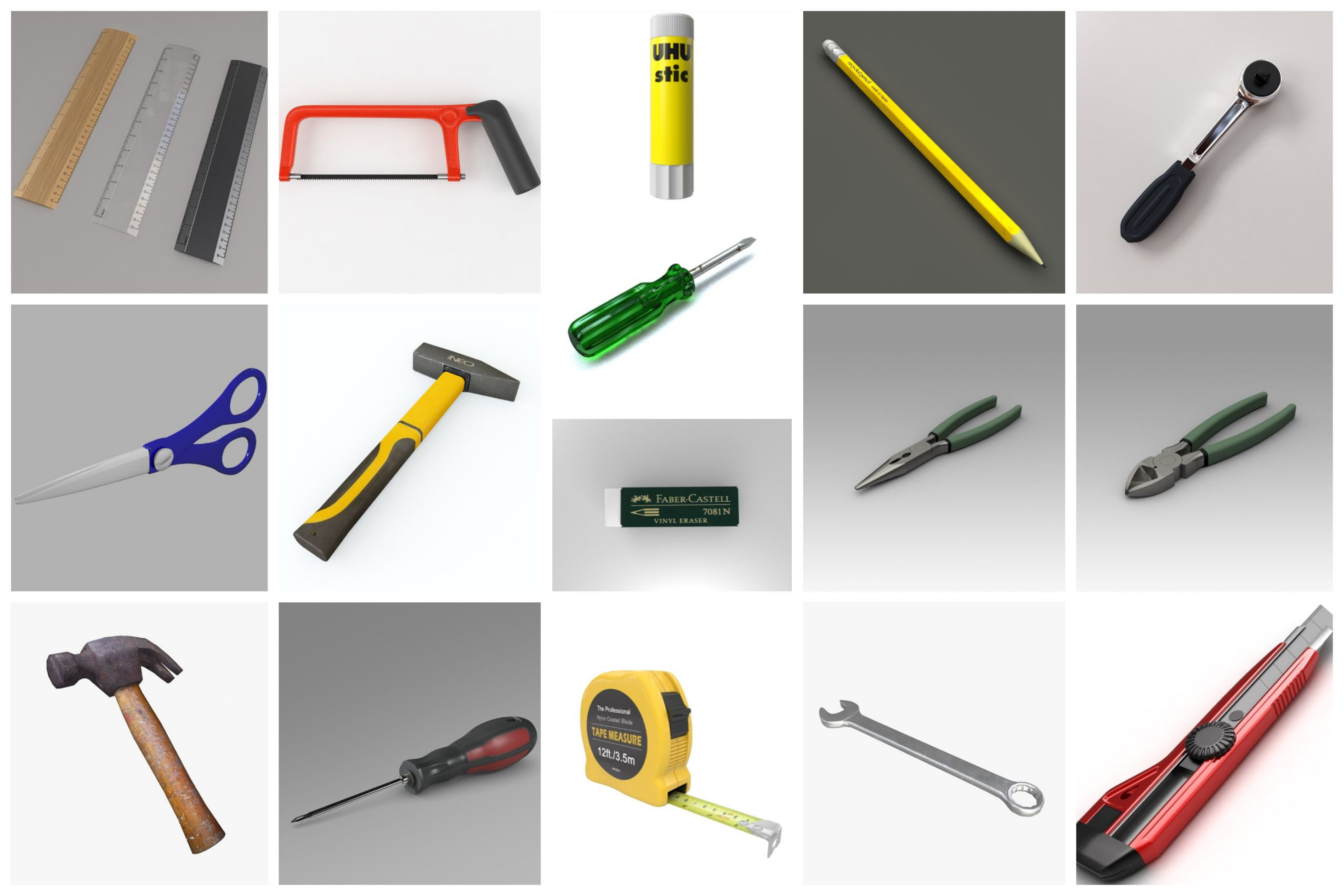}
	\includegraphics[width=0.47\linewidth]{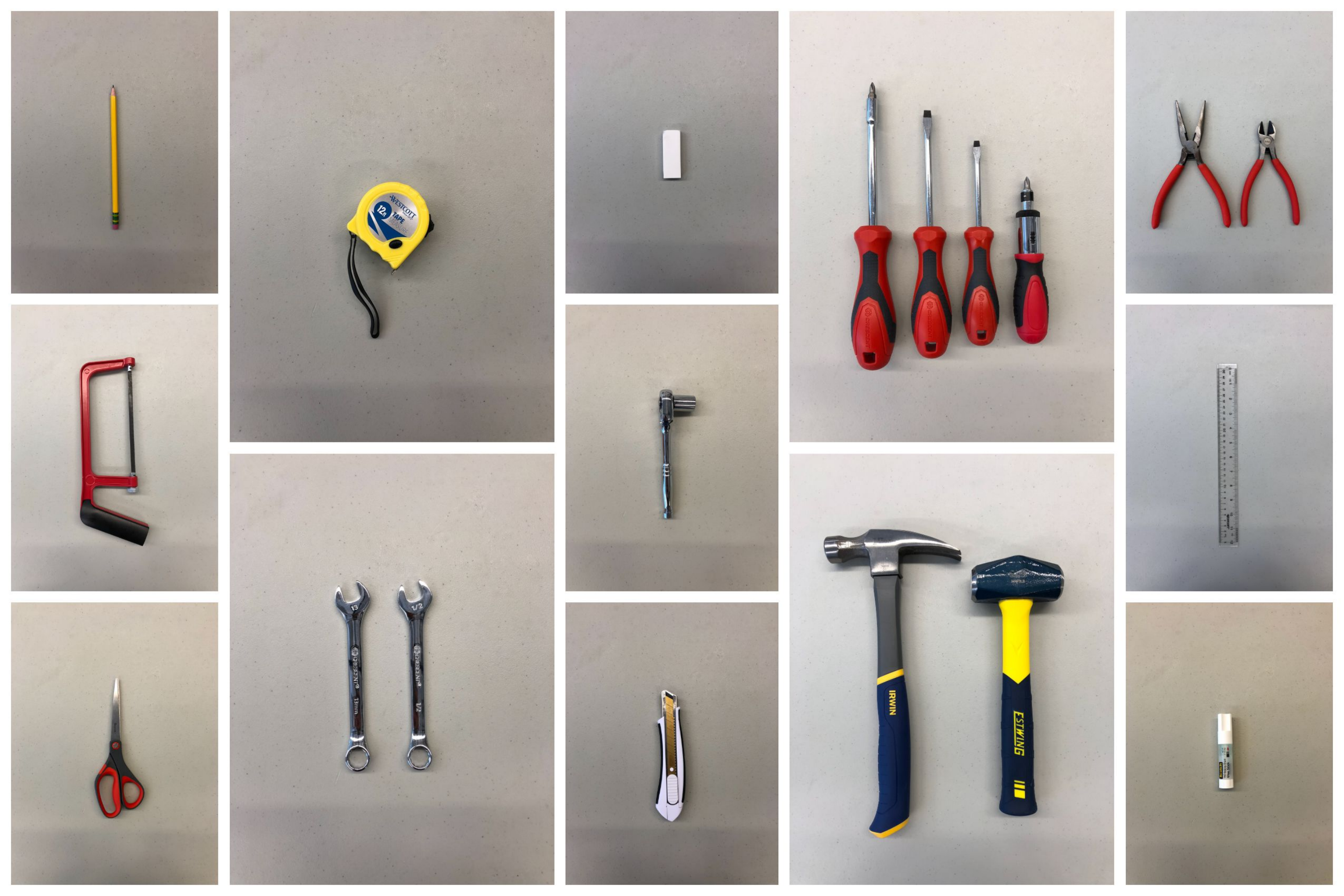}
	\caption{The collection of tools used in BusyHands. Left: Synthetic tool models, Right: Real tools.}
	\label{fig:tools}
\end{figure*}

\section{Related Work}
\label{sec:related}
Semantic segmentation has long been a central pursuit as part of the computer vision research agenda, driven by compelling applications in autonomous navigation, security, image-based search and manufacturing, to name a few. 
In recent years, semantic segmentation research has seen a tremendous boost in offerings of deep convolutional network architectures, marked roughly by Long et al's Fully-Convolutional Networks (FCN) work \cite{LSD15} as the new era of semantic segmentation.
The key insight behind that early work, which still resonates in most of the state-of-the-art contributions of today, is to use a visual feature-extracting network (such as VGG~\cite{KA14}, ResNet~\cite{HZRS15}, or a standalone one) and layer on top of it a decoding and unpooling mechanism to predict a class for each pixel at the original resolution.
In this pattern, we can utilize a rich pre-trained subnetwork with powerful visual representation, proven for example, to work on large-scale image classification problems.
Recent work, such as the flavors of DeepLab \cite{CPKM14,CPKM16,CPSA17}, PSPNet \cite{DBLP:journals/corr/ZhaoSQWJ16} and DenseASPP \cite{Yang_2018_CVPR}, utilize a specialized unpooling device such as the Atrous Spatial Pyramid Pooling (ASPP) feature.

\subsection{Related Segmentation Datasets}
The burst of creativity in semantic segmentation algorithms could not have occurred if not for the equally sharp rise in very large pixel-annotated datasets for segmentation.
With abundance of data, such as PASCAL VOC \cite{EVWW10}, MS COCO \cite{LMBB14}, Cityscapes \cite{CORR16} or ADE20K \cite{ZZPF17}, researchers could build deeper and more influential work, which makes a strong case for building and sharing datasets openly.
Our dataset, on the other hand, offers a far more comprehensive cover of work-tools than any of the aforementioned datasets. 
In Table \ref{tab:instances-compare} we compare the number of pixel-level annotated instances of the objects in our dataset.

Insofar as \textit{hands} are a key element to many useful applications of computer vision, such as egocentric augmented reality or manufacturing, many datasets to segment hands in images were contibuted.
We list a few recent instances in Table \ref{tab:hand-seg-ds}.
However, all of the above mentioned datasets only provide annotation for the hand (up to the wrist), whereas our annotation also provides the \textit{arm} on top of an annotation of the tools in use, while taking great care to mark the hand occlusion from the tools.

\newcommand{\rgbbox}[2]{\fcolorbox{black}[rgb]{#1}{\footnotesize \textbf{$\mathtt{#2}$}}}

\setlength{\tabcolsep}{4pt}
\begin{table}
	\begin{center}
		\begin{tabular}{lllc}
			\toprule
			\# &\textbf{Tool} & \textbf{Assembly Task} & \textbf{RGB} \\
			\hline
			1. &screwdriver  & Tighten or loose screws & \rgbbox{1,1,0}{FFFF00} \\
			2. &wrench & Tighten or loose nuts & \rgbbox{0,1,1}{00FFFF} \\
			3. &pliers & Cut wires  & \rgbbox{1,0,1}{FF00FF}\\
			4. &pencil & Sketch on paper & \rgbbox{0.75,0.75,0.75}{C0C0C0}\\
			5. &eraser & Erase a sketch on paper & \rgbbox{0,0,0.5}{\color{white}000080}\\
			6. &scissors & Cut paper & \rgbbox{ 0.5,0.5,0.5}{\color{white}808080}\\
			7. &cutter & Cut paper & \rgbbox{0.5,0,0}{\color{white}800000} \\
			8. &hammer & Drive nail into wood & \rgbbox{0.5,0.5,0}{\color{white}808000} \\
			9. &ratchet & Tighten or loose nuts & \rgbbox{0,0.5,0}{\color{white}008000}\\
			10. &tape measure & Measure objects & \rgbbox{0.5,0,0.5}{\color{white}800080}\\
			11. &saw & Saw a wooden board & \rgbbox{0,0.5,0.5}{\color{white}008080}\\
			12. &glue & Glue papers & \rgbbox{0.8,0.52,0.25}{\color{white}CD853F}\\
			13. &ruler & Draw line with pencil & \rgbbox{0.27,0.5,0.7}{\color{white}4682B4} \\
			\hdashline
			& hand & & \rgbbox{1,0,0}{FF0000}\\
			& arm & & \rgbbox{0,1,0}{00FF00}\\
			\bottomrule
		\end{tabular}
		\vspace{0pt}
		\caption{Selected tools, their tasks and their mask RGB value (as can be seen in Fig.\ref{fig:samples},\ref{fig:syn_data},\ref{fig:comparison},\ref{fig:results}).}
		\label{table:tools_tasks}
	\end{center}
\end{table}
\setlength{\tabcolsep}{1.4pt}

\section{Constructing the BusyHands Dataset}
\label{sec:dataset}
We chose to deliver two types of image data in BusyHands, real-world and synthetic, so together they can provide a generalized and practical database for semantic segmentation for small-scale assembly works. Real and synthetic data complement each other in number of ways, which we detail in Table \ref{tab:pros-and-cons}.

\begin{figure*}[t]
	\centering
	\includegraphics[width=\linewidth]{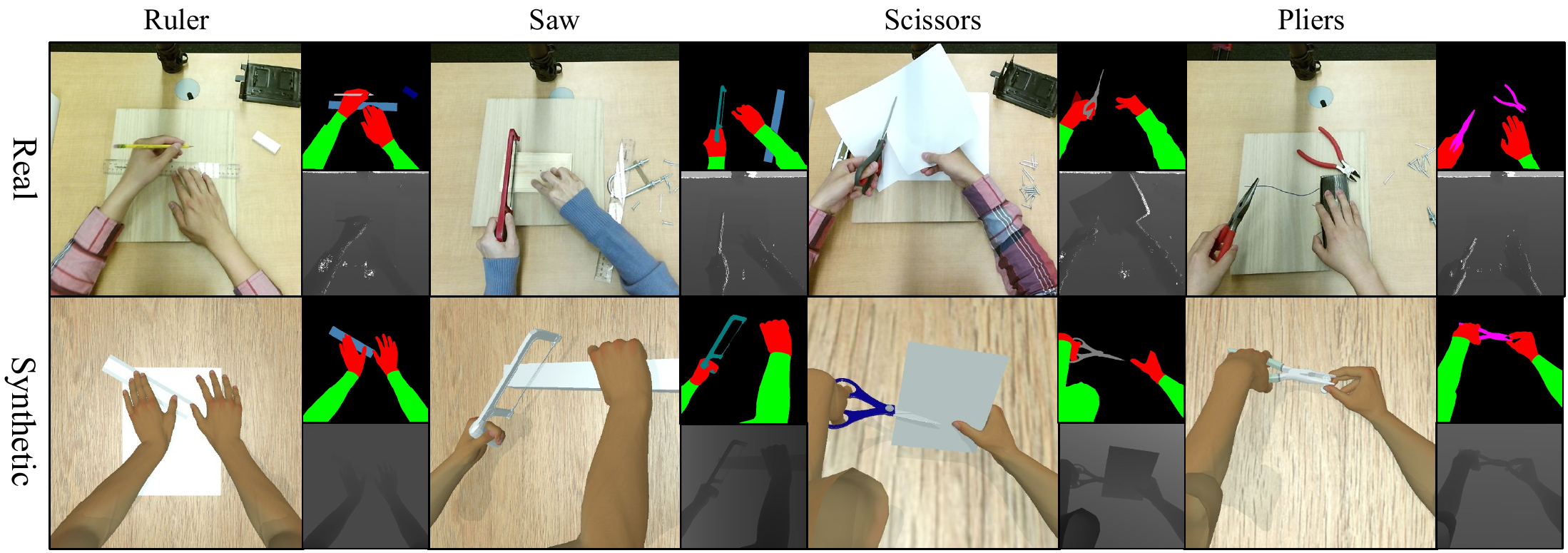}
	\caption{Real vs. synthetic segmentation annotation comparison.}
	\label{fig:comparison}
\end{figure*}

The structure of the dataset is designed following PASCAL~\cite{EVWW10}, which includes color images and segmentation class labels (See Fig.\ref{fig:comparison}). The pixel-value of the segments in the label image ranges from 0 to $N-1$ (where $N = \textrm{\# of classes}$). In addition, we include depth images in our dataset to provide extra information. The work of~\cite{SFCS11,TSLP14}, showed depth images can be extremely useful for understanding human body parts. RGB information is also very hard to generalize properly. In real world situations there is immense color variability, for example shirt, tool, background or skin colors, let alone variation in lighting. Depth images circumvent these problems while the added cost of obtaining them is not high.

\subsection{Tools and Tasks Selection}
We aim to create a dataset for most small-scale assembly works. However, assembly is a widely diverse action with many goals that uses a large class of tools. We chose to focus on common tools that exist in most households and manual assembly pipelines. We used a pre-selected collection of handheld tools (a kit from an established brand) from a home improvement store. Out of the available tools in the kit, we choose 13 common handheld tools listed in Table \ref{table:tools_tasks}.
Pictures of the collection of tools used in our recordings can be seen in Fig.\ref{fig:tools}.

The manual tasks to perform with each tool are derived from the standard function of the tool itself. We staged a small workstation with wooden and paper craft pieces to be used for work pieces, and instructed the ``workers'' to perform simple assembly tasks (see Table \ref{table:tools_tasks}).

\subsection{Real-world Data in BusyHands}

Data was captured using a standard Kinect V2 camera, capturing at 1920 $\times$ 1080 resolution for RGB and 512 $\times$ 424 for depth at 7 FPS. Depth and RGB streams are pixel-aligned using the provided SDK and the camera intrinsic and extrinsic parameters. The frame by frame outputs are demonstrated in Fig.\ref{fig:real_data}.
The camera is mounted above the desk to provide first-person perspective effects. 
This was done to allow our data to be used both for segmentation of images from head-mounted gear as well as top-view cameras in a workbench, which are becoming more and more ubiquitous in the manufacturing world.
During the recording, the real time video output was displayed so that the workers could adjust their postures to avoid excessive occlusion. 
Given the instructions as shown in Table~\ref{table:tools_tasks}, three volunteers were recruited (one female, two males). Skin pigment complexion: one Caucasian, two Asians. 
Multiple tools are allowed to use in one task in order to help complete the work. 
Per each task, the camera started to capture images after the workers began their work, and stopped automatically after recording 150 frames. A total of 39 films were captured, of which 26 were fully annotated with segmentation information.

Annotating the semantic parts in images is a tedious task. We employed Python-LabelMe\footnote{\url{https://github.com/wkentaro/labelme}}, an open source image annotation software based on the original LabelMe project from MIT~\cite{russell2008labelme}, to annotate different semantic parts and assign appropriate labels to them. The results can be seen in Fig.~\ref{fig:samples}. We also show the preprocessed data samples in Fig.~\ref{fig:real_data}. Each sample contains color image, depth image and ground truth.

\begin{figure}[t]
	\centering
	\includegraphics[width=\linewidth]{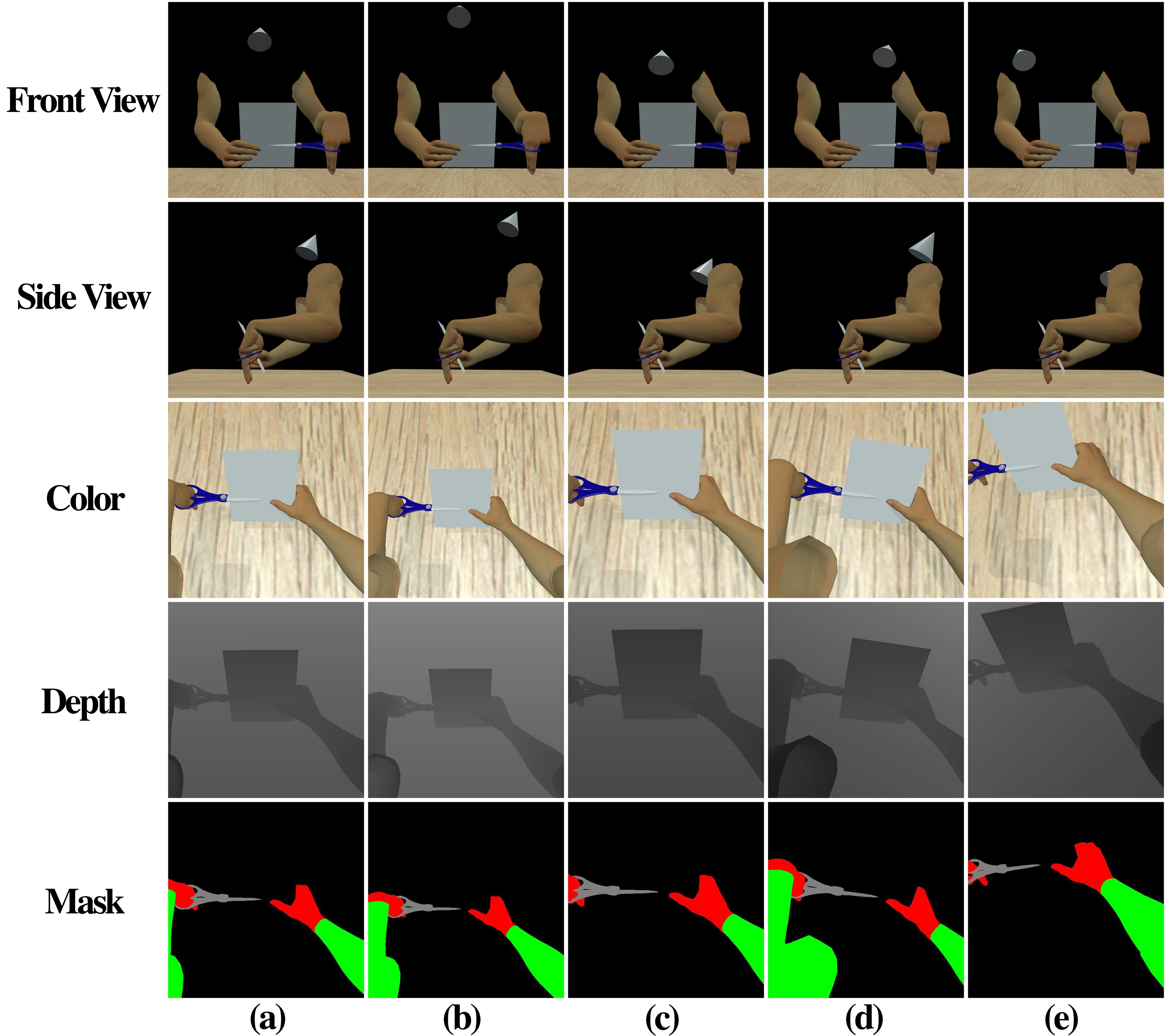}
	\caption{The 3D rendering environment in the synthetic part of BusyHand dataset. For augmentation, we provide 5 camera viewpoints: (a) Center, (b) Up-shift, (c) Down-shift, (d) Left-shift, and (e) Right-shift.}
	\label{fig:syn_data}
\end{figure}

\begin{figure*}[t]
	\centering
	\includegraphics[width=0.49\linewidth]{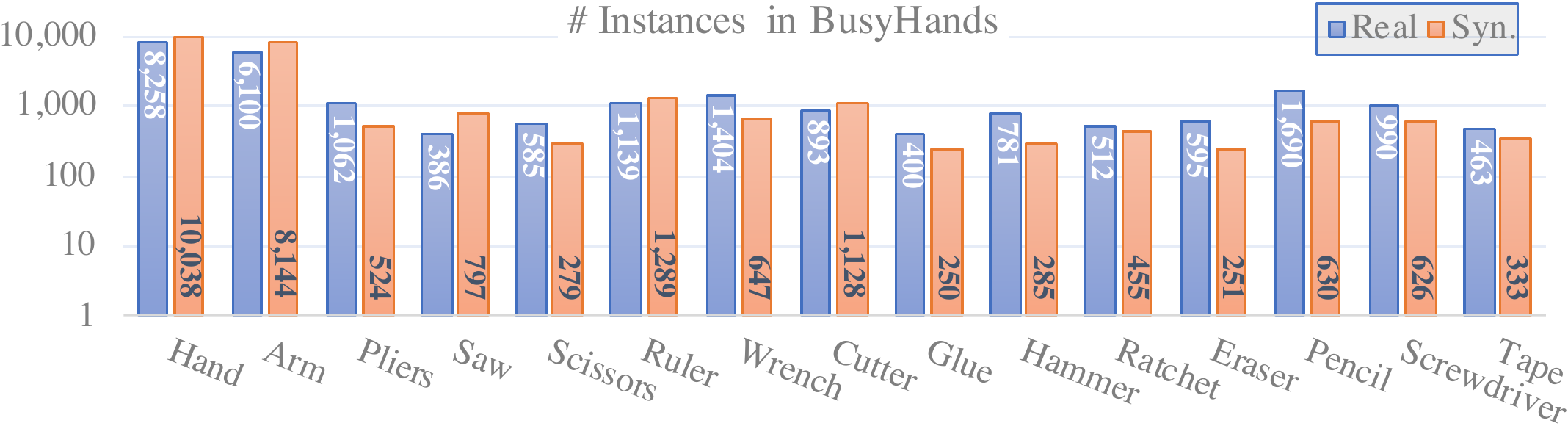}
	\includegraphics[width=0.49\linewidth]{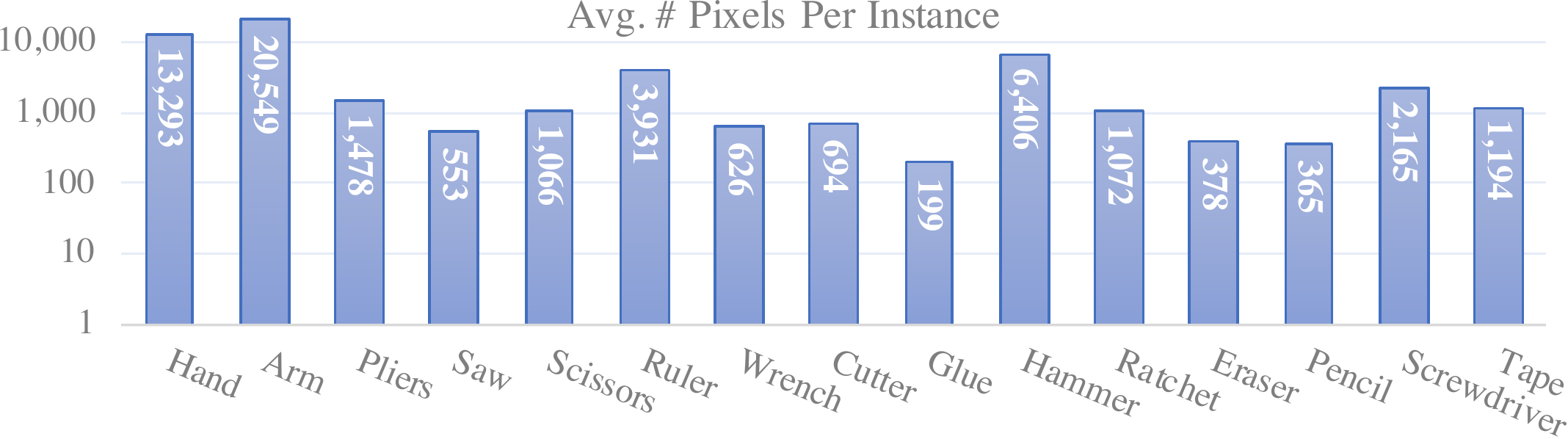}
	\\
	\includegraphics[width=\linewidth]{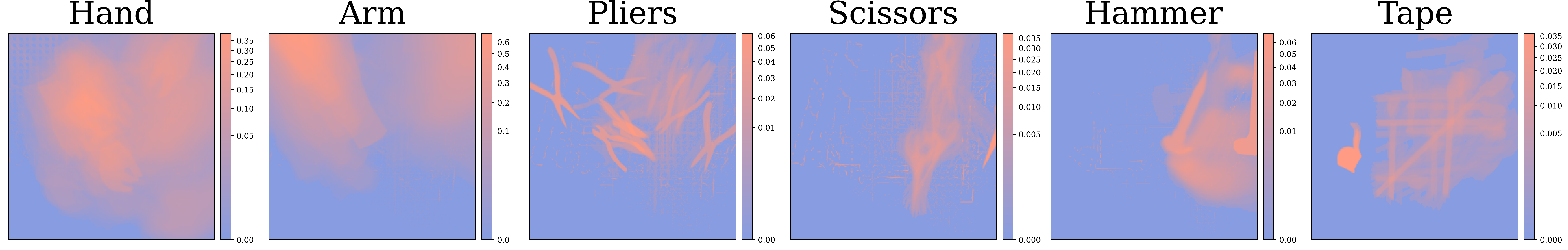}
	\caption{Top: Left: Number of class instances in the BusyHands dataset; Right: Average number of pixels for an instance of each class (e.g. \textit{Hand} instances cover roughly 13,300 pixels on average). Note the logarithmic scale. Bottom: Heatmap illustration of the pixel-position of a few classes in the Real part of the dataset.}
	\label{fig:tools_proportion}
\end{figure*}

\subsection{Synthetic Data in BusyHands}
As mentioned before, to enrich the selection of available data in our dataset and obtain a large number of samples, we adopted using synthetic data. 
To generate realistic data to be on a par with real data, we purchased high quality 3D models of tools (see Fig.~\ref{fig:tools}) as well as a highly realistic pair of hands, and loaded them in the Blender software\footnote{\url{https://www.blender.org/}}. 
All the manual tasks (or instructions) were simulated by creating realistic key-frame animations mimicking human motion by observation.

To increase the generality of the dataset, so it can be applied in various physical environments, we use five camera perspectives in the synthetic dataset. 
As demonstrated in Figure \ref{fig:syn_data}, the cones in the first two rows that represent 5 different camera positions (first-person perspective, move up, move down, move to the left, move to the right) from left to right are rendered in front view (first row) and side view (second row). 
Corresponding color image, depth image and ground truth are given in the bottom three rows.

Unlike real-world captures, annotating semantic parts in a virtual environment is very straightforward. 
In Blender, we unwrapped the meshes of tools, hands, and arms to 2D UV maps, then painted the UV maps using solid colors. 
Each color is one-to-one mapped to one class label in our dataset according to the RGB-codes dictionary (see Table~\ref{table:tools_tasks}). 
Later, we utilize these colors to retrieve corresponding label numbers. 
Given a mapped texture in Blender, the software will output rendered images of RGB and semantic labels for all the designed animation frames automatically. 
A depth map for each frame is easily obtained from Blender by outputting the virtual camera's z-buffer, and is pixel-aligned to the other streams.

\subsection{Dataset Analysis and Comparison}
The real world part of the dataset has 3695 labeled images, while in the synthetic part has 4170 images. 
Instances wise, we have 9505 instances of tools in the real dataset, and 4170 instances of tools in the synthetic parts. 
The proportions of each tool instance for both real data and synthetic data are listed in Fig.~\ref{fig:tools_proportion}.

\section{Semantic Labeling Evaluation}
\label{sec:evaluation}
The BusyHand task involves predicting a pixel level semantic labeling of the image without considering higher level object instance or boundary information.

\subsection{Metrics}

We use a standard metric to evaluate labeling performance. 
The most adopted is the \textit{intersection-over-union} metric $IoU=\frac{TP}{TP+FP+FN}$, where TP, FP, and FN are the numbers of true positive, false positive, and false negative pixels, respectively \cite{EVWW10}. 
We employ an averaging mechanism as is custom, over all classes and then over samples, to achieve the mean intersection over union (mIOU).

\begin{table*}[t]
	\begin{tabular}{lccccccccc}
    \toprule 
     & \multicolumn{9}{c}{Algorithm}  \\
    \cmidrule(lr){2-10} 
		 Train $\rightarrow$ Test 	& AdapNet	& DeepLabV3	& DeepLabV3+	& SegNet	& SegNet-Sk	& FRRN-A	& FRRN-B	& M-UNet	& M-UNet-Sk \\	
         \midrule 
		Rl. $\rightarrow$ Rl.	&0.174	&0.113	&0.139	&0.257	&\textbf{0.336}	&0.316	&0.283	&0.234	&0.22 \\
		Syn. $\rightarrow$ Syn.	&0.714	&0.532	&0.584	&0.782	&0.856	&0.856	&\textbf{0.858	}&0.759	&0.842 \\
		Syn.+Rl. $\rightarrow$ Rl.	&0.291	&0.212	&0.227	&0.328	&0.494	&0.502	&\textbf{0.589	}&0.216	&0.388 \\
		Syn.+Rl. $\rightarrow$ Syn.	&0.623	&0.367	&0.313	&0.591	&0.641	&\textbf{0.776	}&0.763	&0.547	&0.713	\\ 	
        \bottomrule 
	\end{tabular}
	\vspace{0pt}
	\caption{Results of the baseline methods on the BusyHands dataset, in terms of \textit{mIOU}. The first column marks training vs. testing, e.g. `Syn.+Rl. $\rightarrow$ Rl.' means training on both synthetic and real images (training set) and testing only on real images (test set held out). `Sk' indicates the use of skip connections in the network. `M-UNet` is the MobileUNet architecture~\cite{DBLP:journals/corr/HowardZCKWWAA17}.}
	\label{tab:results-miou}
\end{table*}

\begin{figure*}[t]
	\includegraphics[width=\linewidth]{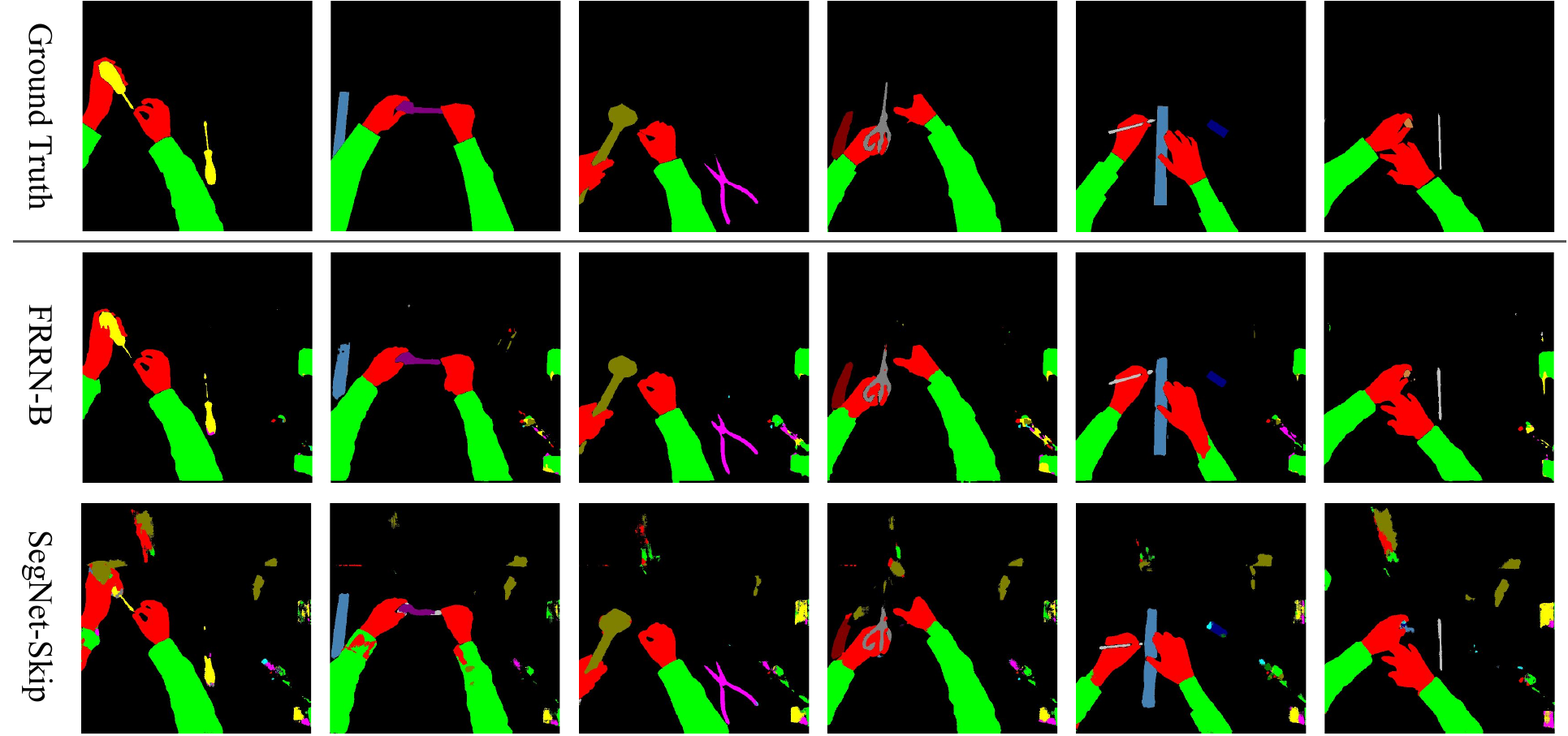}
	\caption{Results of running FRRN-B~\cite{DBLP:journals/corr/PohlenHML16} and SegNet-Skip~\cite{DBLP:journals/corr/BadrinarayananK15} on a number of samples from the Real test dataset. The top row is the ground truth annotation.}
	\label{fig:results}
\end{figure*}

\subsection{Evaluated Segmentation Methods}
\label{sec:methods}
We experimented with the following semantic segmentation algorithms, from the latest literature:

\begin{itemize}
\item \textbf{Encoder-Decoder SegNet \cite{DBLP:journals/corr/BadrinarayananK15}}. This network uses a VGG-style encoder-decoder, where the upsampling in the decoder is done using transposed convolutions.
In addition, we also used a version that employs additive skip connections from  encoder to decoder.

\item \textbf{Mobile UNet for Semantic Segmentation \cite{DBLP:journals/corr/HowardZCKWWAA17}}. Combining the ideas of MobileNets Depthwise Separable Convolutions with UNet results in a low-parameter semantic segmentation model. In this architecture we also have a flavor with skip connections.

\item \textbf{Full-Resolution Residual Networks (FRRN) \cite{DBLP:journals/corr/PohlenHML16}}. Combines multi-scale context with pixel-level accuracy by using two processing streams within the network. The residual stream carries information at the full image resolution, enabling precise adherence to segment boundaries. The pooling stream undergoes a sequence of pooling operations to obtain robust features for recognition. The two streams are coupled at the full image resolution using residuals. 

\item \textbf{AdapNet \cite{valada2017adapnet}}. Modifies the ResNet50 architecture by performing the lower resolution processing using a multi-scale strategy with atrous convolutions. We use a slightly modified version using bilinear upscaling instead of transposed convolutions.

\item \textbf{DeepLabV3 \cite{CPSA17} and DeepLabV3+ \cite{CZPS18}}. Uses Atrous Spatial Pyramid Pooling to capture multi-scale context by using multiple atrous rates. This creates a large receptive field.
The DeepLabV3+ network adds a Decoder module on top of the regular DeepLabV3 model.
\end{itemize}

All algorithms were implemented with the Tensorflow package \cite{abadi2016tensorflow}, forking the Semantic Segmentation Suite project \cite{semanticsegmentationsuite} to which we made several adjustments.

\subsection{Evaluation Results}

The results of training and testing with the selected evaluation methods (listed in \S\ref{sec:methods}) are given in Table \ref{tab:results-miou}. 
We notice that the full-resolution residual networks (FRRNs) are mostly superior under all categories, followed by the SegNet with skip connections. 
In Figure \ref{fig:results} we show example results on the Real test set with FRRN-B and SegNet-Skip (additional results are available as supplementary material).
The results indicate that while segmenting the arms, hands and tools is done quite well, there is a significant amount of noise from random objects on the table that classify as tools.
Some post processing cleanup on the segmentation result, in particular blob geometry analysis (which we did not attempt), could potentially alleviate the level of noise.

Another insight is that the existence of synthetic data dramatically increases the power of the learners in accuracy over Real data. 
In the case of FRRN-A, for example, mIOU over the Real test set shot up from 0.336 when training just with Real images up to 0.502 when using also synthetic data for training.
In fact only in the case of MobileUNet the performance dropped when including synthetic data, otherwise it increased performance by up to \%80 throughout.

\section{Conclusions}
\label{sec:conclusion}
We contribute BusyHands - a high-quality fully annotated dataset for semantic segmentation with both real and synthetic image data. 
We also present an evaluation of numerous leading segmentation algorithms on our dataset as a baseline for other researchers.
We release all of the data for general access of the computer vision community  at \href{http://hi.cs.stonybrook.edu/busyhands}{http://hi.cs.stonybrook.edu/busyhands}.
This, we hope, will allow to create better image segmentation algorithms, which will even further advance computer vision research on scenes of manual assembly operations.

\section*{Acknowledgments}
We would like to thank Nvidia for their generous donation of a Titan Xp and Quadro P5000 GPUs, which were used in this project.
We thank the dataset annotators: Sirisha Mandali, 
Venkata Divya Kootagaram, as well as Fan Wang, Xiaoling Hu.

\bibliographystyle{splncs}
\bibliography{egbib}

\end{document}